%% file: main.tex
\documentclass[10pt,twocolumn,letterpaper]{article}

 \usepackage[pagenumbers]{cvpr} 

\input{preamble}

\definecolor{cvprblue}{rgb}{0.21,0.49,0.74}
\usepackage[pagebackref,breaklinks,colorlinks,allcolors=cvprblue]{hyperref}

\def\confName{CVPR}
\def\confYear{2027}

\title{ActTrail: Action-Faithful Video World Modeling with Predicted Point Trajectories}

\author{
Ao Tian$^{*}$ \qquad
Yibo Qiu$^{*}$ \qquad
Shu'ang Sun \qquad
Mingzhai Sun$^{\dagger}$\\
University of Science and Technology of China\\
{\small $^{*}$These authors share first authorship.
\qquad $^{\dagger}$Corresponding author.}
}
\usepackage{cuted}
\begin{document}

\maketitle
\input{fig/teaser}

\input{sec/0_abstract}  
\input{sec/1_intro}
\input{sec/2_related_work}
\input{sec/3_method}
\input{sec/4_experiments}
\input{sec/5_conclusion}
\input{sec/6_llm_usage}
{
    \small
    \bibliographystyle{ieeenat_fullname}
    \bibliography{main}
}


\end{document}


\maketitlesupplementary

\section{Datasets and Evaluation Protocols}
\label{sec:supp-data-eval}

\subsection{Splits, sampling, and action alignment}
\label{sec:supp-data}

We evaluate 12-step action-conditioned video prediction on BridgeData V2
and LIBERO. For each test clip, all methods receive the same recorded initial
frame $I_0$ and sequence of seven-dimensional actions $a_1,\ldots,a_{12}$.
The targets are the next 12 frames $I_1,\ldots,I_{12}$. For a clip starting
at source frame $s$, $I_t$ is frame $s+t$ and $a_t$ is row $s+t-1$ of the
conditioning sequence, aligned with the transition from $I_{t-1}$ to $I_t$.
Source indices start at zero.

\noindent\textbf{BridgeData V2.}
We use the BridgeData V2 release distributed by IRASim and remove
exact duplicates based on decoded RGB content. Using a fixed selection
rule, we obtain disjoint sets of 2,000 training, 64 development, and
128 test trajectories. We extract every complete 13-frame window from
the training trajectories with stride one, yielding 41,142 clips.
For development and testing, we deterministically select one window
per trajectory. Each clip contains consecutive frames from camera~0
at 3~Hz and spans 4~s from $I_0$ to $I_{12}$.

We construct the action condition from the end-effector poses and gripper
states associated with adjacent released frames. Let $p_t$, $R_t$, and $g_t$
denote end-effector position, orientation, and continuous gripper state at
frame $t$. We express the translation and relative rotation in the previous
end-effector frame as
\[
\begin{aligned}
  \Delta p_t &= R_{t-1}^{\top}(p_t-p_{t-1}),\\
  \Delta\theta_t &= \operatorname{Euler}_{xyz}(R_{t-1}^{\top}R_t),\\
  a_t &= (20\,\Delta p_t,\;20\,\Delta\theta_t,\;g_t).
\end{aligned}
\]
Translation is in meters before scaling, following the upstream Bridge
collection convention. The released orientation angles are in radians and
use extrinsic XYZ rotations, with
$R=R_z(\gamma)R_y(\beta)R_x(\alpha)$ for angles $(\alpha,\beta,\gamma)$.
We decompose the relative rotation using the same convention and store the
three angles in $x,y,z$ order. Each action uses only its corresponding pair
of released frames, with the target-frame gripper state as its seventh entry.

\noindent\textbf{LIBERO.}
For each of the 40 tasks, we split complete demonstration episodes into 40
training, five development, and five test episodes. This yields 1,600, 200,
and 200 episodes across the three splits. For a test episode of $T$ frames,
let $L=T-13$. We extract four clips starting at $0$, $\lfloor L/3\rfloor$,
$\lfloor 2L/3\rfloor$, and $L$. These starts are distinct for every test
episode, yielding 800 test clips. We use consecutive observations from the
agent-view camera at one frame per 20~Hz control step. The interval from
$I_0$ to $I_{12}$ is 0.6~s.

LIBERO records the observation before executing the control command. Action
row $s+t-1$ therefore contains the seven-dimensional command executed after
observing $I_{t-1}$ and leading to $I_t$. Data regeneration skips no-op
commands before simulation. We read the retained commands in execution
order, one per transition. DreamDojo and ActTrail place each command in the
designated slots of their conditioning interfaces. IRASim casts the LIBERO
control values to BF16 at inference with the same action-to-frame alignment.

\noindent\textbf{Image preprocessing and evaluation coordinates.}
All methods retain the full field of view of the initial frame.
DreamDojo, Ctrl-World, and ActTrail use the native RGB dimensions of
$480\times640$ on Bridge and $256\times256$ on LIBERO
(height $\times$ width). IRASim and Vid2World retain the input dimensions
of their upstream models. For both methods on each dataset, inference uses
the same resizing, image extent, and padding as fine-tuning.

IRASim resizes Bridge frames to $256\times320$ using bilinear
interpolation and horizontally zero-pads LIBERO frames to the same
size. Vid2World uses bicubic interpolation to resize Bridge and LIBERO
frames to $320\times427$ and $320\times320$, respectively, then
zero-pads both horizontally to $320\times512$. Before evaluation,
we remove any padding and, where needed, resize predictions to the
native resolution using the same interpolation method as for the input.
We measure motion in native pixels and compare RGB values at
corresponding image locations.

\subsection{Motion and dynamic-region reconstruction metrics}
\label{sec:supp-metrics}

\noindent\textbf{Query points and track correspondence.}
We use SAM3 to segment the robot in the recorded initial frame and select
query points deterministically at least five pixels inside the mask. On
Bridge, we select up to 16 points across the full mask interior. A
reference track must be valid in the initial frame and in at least 80\% of
the 13 frames. We retain clips with at least eight such reference tracks
for motion evaluation. On LIBERO, we select 16 interior points in the
upper half of the foreground bounding box.

CoWTracker tracks the same initial queries separately in the recorded video
and each generated video. We match tracks by their initial query indices
and map coordinates from the tracker's internal $336\times560$ resolution
back to native image coordinates. A query is valid in a frame if its
coordinates, visibility, and confidence are finite, both visibility and
confidence are at least 0.5, and its coordinates fall inside the image.
Generated tracks use per-query, per-frame validity. The 80\% track-level
requirement applies only to the Bridge reference tracks.

\noindent\textbf{Motion errors.}
Let $\mathbf{x}_{tq}$ and $\hat{\mathbf{x}}^m_{tq}$ denote the recorded
and generated 2D positions of query $q$ at frame $t$ for method $m$.
For a frame pair $(i,j)$, define the displacement error
\[
e_m(i,j,q)=\bigl\|
(\hat{\mathbf{x}}^m_{jq}-\hat{\mathbf{x}}^m_{iq})
-(\mathbf{x}_{jq}-\mathbf{x}_{iq})\bigr\|_2.
\]
For each clip and comparison group, $Q_{i,j}$ contains the queries valid in
both frames $i$ and $j$ in the recorded video and in every generated video
in the group. The three per-clip motion scores are
\[
\begin{aligned}
\mathrm{Step}_m &=
\frac{\sum_{t=1}^{12}\sum_{q\in Q_{t-1,t}} e_m(t-1,t,q)}
     {\sum_{t=1}^{12}|Q_{t-1,t}|},\\[2pt]
\mathrm{Mean\,cum.}_m &=
\frac{\sum_{t=1}^{12}\sum_{q\in Q_{0,t}} e_m(0,t,q)}
     {\sum_{t=1}^{12}|Q_{0,t}|},\\[2pt]
\mathrm{End}_m &=
\frac{1}{|Q_{0,12}|}\sum_{q\in Q_{0,12}}e_m(0,12,q).
\end{aligned}
\]
Step measures displacement error between adjacent frames in pixels per
step. Mean cum.\ measures displacement error relative to the initial frame
across all valid future query--time pairs. End measures displacement error
from the initial frame to step~12. Both are in pixels. Step and Mean cum.\
weight all contributing query--time pairs equally within a clip. End
requires the same query to be valid at both endpoints and permits invalid
intermediate frames. A clip with a zero denominator is omitted from the
corresponding metric.

\noindent\textbf{Dynamic-region reconstruction.}
All methods share masks derived from the recorded video. We use SEA-RAFT-S
to estimate bidirectional optical flow between each pair of recorded frames
$(I_{t-1},I_t)$. For a target pixel $u$ in $I_t$, let $b=b_t(u)$ be its
backward flow and let $f=f_t(u+b)$ be the forward flow sampled bilinearly
at the backward-mapped position. The pixel belongs to the dynamic region
$D_t$ if $u+b$ is inside the image, $\|b\|_2>1$ pixel, and
\[
\|b+f\|_2\leq\max\!\left(1,\,0.05\max(\|b\|_2,\|f\|_2)\right).
\]
We erode $D_t$ three times with a $3\times3$ rectangular kernel to obtain
the dynamic interior $M_t$. For each future frame with at least 128
interior pixels, we compute
\[
\ell^m_t=
\frac{1}{3|M_t|}\sum_{u\in M_t}\sum_{c\in\{R,G,B\}}
\left|\hat I^m_t(u,c)-I_t(u,c)\right|.
\]
Dyn.~L1 is the arithmetic mean of $\ell_t^m$ over these qualifying frames
among $t=1,\ldots,12$. RGB intensities are on a 0--255 scale. Clips with
no qualifying frame are omitted from this metric. Boundary L1 uses the
inner boundary $D_t\setminus M_t$ and includes each frame with a nonempty
boundary.

\noindent\textbf{Aggregation across clips.}
On Bridge, each clip with a valid score receives equal weight for that
metric. On LIBERO, we first average valid clip scores within each episode,
then average eligible episodes within each task, and finally average the
40 task means with equal weight.

\subsection{Common valid tracks and test coverage}
\label{sec:supp-coverage}

For each motion metric, all methods in a comparison group are evaluated on
the same query points at the same frame pairs. Dyn.~L1 uses the shared
reference flow masks independently of the trajectory intersection.
Table~\ref{tab:supp-clip-coverage} gives clip counts for each main-paper
comparison.

\begin{table}[tbp]
\centering
\caption{\textbf{Test clips contributing to each metric.}
The first column identifies the main-paper table and comparison. $N$ is
the total number of test clips. Main-paper Table~1 compares the five
methods listed in Table~\ref{tab:supp-end-coverage}. The remaining rows
are Bridge comparisons. Table~5 compares B-Mod, Feature Warping, and
ActTrail. Table~6 compares Full with each named variant separately.
A dash denotes a metric omitted from the corresponding main-paper table.}
\label{tab:supp-clip-coverage}
\small
\setlength{\tabcolsep}{2pt}
\begin{tabular*}{\linewidth}{@{\extracolsep{\fill}}lrrrrr@{}}
\toprule
Group & $N$ & Step & Mean cum. & End & Dyn.~L1 \\
\midrule
1, Bridge & 128 & 120 & 120 & 75 & 128 \\
1, LIBERO & 800 & 791 & 791 & 775 & 758 \\
5, Fusion & 128 & 122 & --- & 117 & 128 \\
6, PShared & 128 & 122 & --- & 120 & 128 \\
6, Joint & 128 & 122 & --- & 121 & 128 \\
\bottomrule
\end{tabular*}
\end{table}

\noindent\textbf{Endpoint coverage on Bridge.}
Of the 128 Bridge test clips, 120 contain a query valid across the
reference and all five methods at frame~0 and at least one future frame,
so they contribute to Mean cum.\ Of these, 75 also contain a common valid
query at frames~0 and~12 and contribute to End. In the other 45 clips,
initial-frame displacement can be evaluated only at earlier prediction
steps.

\begin{table}[tbp]
\centering
\caption{\textbf{Endpoint coverage on Bridge for main-paper Table~1.}
Each row intersects the indicated method or methods with the recorded
reference. A clip contributes if at least one query is valid at both
endpoints. Each contributing initial query counts as one track within
its clip. The last row gives the intersection used to score all five
methods in the main table.}
\label{tab:supp-end-coverage}
\small
\setlength{\tabcolsep}{3pt}
\begin{tabular*}{\linewidth}{@{\extracolsep{\fill}}lrrr@{}}
\toprule
Methods & Valid clips & Coverage & Tracks \\
\midrule
DreamDojo & 121/128 & 94.53\% & 1,683 \\
IRASim & 94/128 & 73.44\% & 934 \\
Vid2World & 118/128 & 92.19\% & 1,505 \\
Ctrl-World & 99/128 & 77.34\% & 1,044 \\
ActTrail & 121/128 & 94.53\% & 1,654 \\
\midrule
All five & 75/128 & 58.59\% & 630 \\
\bottomrule
\end{tabular*}
\end{table}

Table~\ref{tab:supp-end-coverage} shows why individual and shared coverage
differ. Six of the 128 clips fail the reference-track filter. Among the
clips that pass, 40 lack a valid endpoint for at least one method. In
another seven clips, each method has a valid endpoint, but no single query
is valid for all five. The resulting intersection contains 630 tracks in
75 clips. Averaging endpoint displacement errors first within each clip
and then equally across these 75 clips gives the ActTrail End score of
27.11~pixels reported in main-paper Table~1.

\noindent\textbf{Full across comparisons.}
Main-paper Tables~1, 5, and~6 reuse the same ActTrail (Full) predictions
for the same 128 Bridge test clips. Each comparison takes the intersection
over its own set of methods. Both the eligible clips and the queries
within a given clip can therefore change. Full's End is 27.11~pixels
in the five-method comparison (75 clips) and 35.16~pixels in the
three-way fusion comparison (117 clips). In the two comparisons in
Table~6, its End is 37.05~pixels against PShared (120 clips) and
36.62~pixels against Joint (121 clips). Error differences within each
comparison thus refer to the same endpoint tracks. The endpoint comparison in Table~\ref{tab:supp-stage-motion} uses
points valid across the recorded reference and all three trajectory
estimates (121 clips), while Table~\ref{tab:supp-fusion} uses points
valid across the reference and the three fusion methods (117 clips).

\section{Implementation and Training Details}\label{sec:sup-implementation}

ActTrail predicts the 2D trajectories of first-frame points over the next
12 action steps and uses these trajectories to guide video generation.
During training, cumulative tracks and adjacent-frame optical flow
supervise the motion of the same first-frame point. During generation,
four action steps share a latent feature map. The generator reads and
writes features at each step's predicted position before combining the
four updates in their shared temporal slot.

\subsection{Trajectory Predictor and Motion Supervision}
\label{sec:sup-trajectory-predictor}

\noindent\textbf{Two motion targets for each first-frame point.}
Given the initial frame $I_0$ and a sequence of seven-dimensional robot
actions $a_{1:12}$, Stage~1 predicts dense motion on a $60\times80$
grid in the initial frame. For a grid point $q$, $P_t(q)$ denotes its
cumulative 2D displacement from frame~0 to frame~$t$, in grid units.
With $P_0(q)=0$, the displacement of the same point from frame~$t-1$
to frame~$t$ is $V_t(q)=P_t(q)-P_{t-1}(q)$. We supervise $P_t$ with
tracks anchored in the first frame and $V_t$ with adjacent-frame flow.

We construct both targets on working images of size $480\times640$.
We report image dimensions as height by width. A zero-based grid index
$q=(x,y)$ has working-image coordinates
$c=8(q+\tfrac12)-\tfrac12$. AllTracker~V2 estimates cumulative
displacements from the first frame of each ground-truth 13-frame clip.
At each step, we bilinearly sample its displacement field at the
fixed location $c$ and divide by 8 to obtain the target
$P_t^{\mathrm{AT}}(q)$ in grid units. DPFlow estimates the forward
flow $F_t$ from $I_{t-1}$ to $I_t$. Since $F_t$ is defined on
$I_{t-1}$, we sample it at the point's teacher-estimated position in
that frame. The resulting stepwise target is
\begin{equation}
 V_t^{\mathrm{DP}}(q)
 =\tfrac18 F_t\!\left(c+8P_{t-1}^{\mathrm{AT}}(q)\right).
 \label{eq:sup-flow-target}
\end{equation}
We evaluate $F_t$ by bilinear interpolation. The first step samples
flow at $c$. From the second step onward, the sampling location
follows the teacher track, keeping both targets aligned to the same
first-frame point.

Both losses use the same valid point--time pairs. The AllTracker
endpoint must lie inside the image, and the point must have visibility
scores of at least $0.5$ in both the current and previous frames.
We treat the first frame as visible. The DPFlow sampling location
and flow endpoint must lie inside the image, and the path obtained
by composing successive flow fields must remain valid through that
step. Both targets and the composed displacement must be finite.
We apply $\rho(e)=\sqrt{\lVert e\rVert_2^2+10^{-6}}$ to the
cumulative error $P_t-P_t^{\mathrm{AT}}$ and the stepwise error
$V_t-V_t^{\mathrm{DP}}$, and average each loss over all valid
point--time pairs in the batch. On both BridgeData~V2 (Bridge) and
LIBERO, we optimize $\mathcal{L}_{\mathrm{flow}}+\lambda\mathcal{L}_P$
with $\lambda\approx0.1$. The trajectory-only configuration in
Table~4 of the main paper optimizes $\mathcal{L}_P$ with unit weight.
The jointly supervised DINOv2 configuration in that table uses the
same trajectory predictor as the Bridge ActTrail model in Table~1 of
the main paper.

\noindent\textbf{Aligning first-frame features to the trajectory grid.}
Bridge uses its native $480\times640$ initial frame. For LIBERO,
we bilinearly resize the native $256\times256$ initial frame to
$480\times640$, round the values, and convert back to 8-bit RGB.
One trajectory-grid unit thus corresponds to 8 native pixels on
Bridge. On LIBERO, it corresponds to $256/80$ native pixels
horizontally and $256/60$ vertically. These transformations apply
to the trajectory predictor. Section~\ref{sec:supp-data-eval}
gives the input sizes of the video generators and the coordinate
systems used for evaluation.

For both datasets, we bilinearly resize the full working image to
$256\times464$, normalize it using the ImageNet channel statistics,
and pass it to a frozen DINOv2-L/14 encoder with 24 layers and a
feature width of 1,024. Its unpadded patch projection has kernel
size and stride 14, producing $18\times33$ spatial tokens that
cover the first 252 rows and 462 columns. With pixel centers at
integer coordinates, a working-image point $c$ maps to location $d$
in the encoder input and continuous patch index $p$ according to
\begin{equation}
\begin{aligned}
 d &= (c+\tfrac12)\odot(464/640,\,256/480)-\tfrac12,\\
 p &= \bigl(d-(6.5,6.5)\bigr)/14.
\end{aligned}
\label{eq:sup-patch-grid}
\end{equation}
The vectors $c$, $d$, and $p$ use $(x,y)$ order, so $464/640$ and
$256/480$ are the horizontal and vertical scale factors, respectively.
Here $\odot$ denotes elementwise multiplication and $(6.5,6.5)$
is the center of the first patch. For image resizing, we set
\texttt{align\_corners=False} and disable antialiasing. After
removing the CLS token, we bilinearly sample patch features at $p$
with \texttt{align\_corners=True}. Along either axis, queries beyond
the outermost patch center use the corresponding border feature.
A $1\times1$ convolution followed by 32-group GroupNorm and SiLU projects
the resulting $60\times80$ feature map to 256 channels.

\noindent\textbf{Action-conditioned trajectory predictor.}
At each step, we insert the seven action components into fixed
positions in a 384-dimensional conditioning vector, encode it with a
$384\!\to\!256\!\to\!256$ MLP, and add a learned step embedding.
A two-layer causal Transformer encodes one action representation
per step, with four attention heads and a feed-forward width of 512
in each layer. The representation at step~$t$ depends only on
$a_{1:t}$.

All steps share a spatial U-Net. It takes the same first-frame
feature map at every step and uses FiLM to apply the corresponding
action representation. The U-Net has 64, 96, and 128 channels at
its three scales. Two downsampling stages, two bilinear upsampling
stages, and skip connections return the output to the trajectory-grid
resolution. Each of its five $3\times3$ convolutional units uses
eight-group GroupNorm, action-conditioned FiLM, and SiLU. A causal
temporal convolution with kernel width three, left padding, and
zero-initialized weights mixes features across neighboring steps
through a residual connection. All steps share a $1\times1$
displacement head. Its output has shape
$B\times12\times2\times60\times80$ and contains the cumulative
displacement fields.

AllTracker~V2 and DPFlow provide cumulative-track and adjacent-frame
flow references, respectively, for Stage~1. At inference, ActTrail
predicts trajectories from the initial frame and actions alone,
without either teacher, future ground-truth frames, target
trajectories, or optical flow. We evaluate motion and reconstruction in the generated videos
using the metrics defined in Section~\ref{sec:supp-metrics}.

\subsection{Trajectory Fusion in the Video Generator}
\label{sec:sup-trajectory-fusion}

\noindent\textbf{Four action steps in one temporal slot.}
Stage~2 uses the DreamDojo-2B action-conditioned video DiT and a frozen Wan2.1 video VAE.
The VAE encodes 13 frames into four latent temporal slots, one for the initial frame and one for each group of future steps (1--4, 5--8, and 9--12).
For steps 1--4, each trajectory specifies four positions on the same feature map.
The trajectory branch reads a feature at each position, applies attention along the trajectory, and writes each node's output back at its own position.
It then combines the four spatial update maps into a single residual for the slot.

\noindent\textbf{Feature sampling and attention along each trajectory.}
Each initial-frame video token defines a trajectory with 13 nodes.
We map the token center to the $60\times80$ Stage~1 grid, interpolate all 12 cumulative displacements at that fixed initial position, and map the resulting endpoints back to the video-token grid using pixel-center coordinates.
The video VAE compresses each spatial dimension by a factor of eight and produces 16 latent channels. The DiT uses $1\times2\times2$ spatiotemporal patches.
The spatial grids for latents and video tokens are therefore $60\times80$ and $30\times40$ on Bridge, and $32\times32$ and $16\times16$ on LIBERO, respectively.

All 28 DiT blocks include a trajectory branch with a feature width of 2,048 and 16 attention heads.
Each branch samples node features from the normalized, conditionally modulated feature map immediately before self-attention.
The initial node reads from its original position in slot~0. Each future node bilinearly samples its temporal slot at its predicted position.
A channel-wise affine transformation depends on the node's step within its slot, with separate parameters for the initial node.
Attention operates within each trajectory. Its queries and keys receive 3D rotary positional embeddings with coordinates $(t/4,y_t,x_t)$, where $(x_t,y_t)$ is the node position on the video-token grid.

A query can attend to valid nodes up to the end of its own slot.
Thus, queries at steps 1--4 can attend to valid nodes 0--4, including later steps in the same slot.
Queries at steps 5--8 and 9--12 can attend to valid nodes 0--8 and 0--12, respectively.
The initial node attends only to itself and supplies a key and value for future queries.

\noindent\textbf{Stepwise spatial write-back and within-slot aggregation.}
At each action step, we splat each valid node's attention output onto the video tokens surrounding its predicted endpoint using bilinear weights.
When multiple trajectory points contribute to the same destination token, we sum their weighted features and divide by the sum of the interpolation weights at that token for that step.
The update is zero wherever the weight sum is at most $10^{-6}$.
For nodes with non-finite coordinates or positions outside the range of token centers, we set the sampled features and attention outputs to zero, mask their keys, and omit their write-back contributions.

Let $S_t^{(\ell)}$ denote the spatial update map in DiT block $\ell$ after splatting and normalization at step $t$.
For the first future slot, the additional residual is
\begin{equation}
\Delta H_1^{(\ell)}
= W_{\mathrm{out}}^{(\ell)}
\left(\sum_{j=1}^{4} w_j^{(\ell)}\odot S_j^{(\ell)}\right),
\label{eq:sup-slot-residual}
\end{equation}
where $w_j^{(\ell)}$ is a learned vector of channel-wise weights for the $j$th step within a slot.
The three future slots in a block share the four weight vectors. For example, steps 1, 5, and 9 all use $w_1^{(\ell)}$.
Each vector is shared across spatial locations, initialized near 0.25, and used directly in the sum without a softmax.
The bias-free output projection $W_{\mathrm{out}}^{(\ell)}$ is initialized to zero, so the branch contributes zero residual at the start of training.
We add the trajectory residual for each future slot after the DiT's gated self-attention residual update and before conditional cross-attention and the feed-forward network.
Slot~0 supplies features to the trajectory branch and always receives zero additional residual.

\subsection{Training and Inference Settings}
\label{sec:sup-training}

\noindent\textbf{Two-stage training.}
We train a separate trajectory predictor for each dataset. Stage~1 freezes
DINOv2 and trains the visual projection, action encoder, and displacement
predictor. Stage~2 freezes the dataset-specific trajectory predictor and the
video VAE. The video DiT and its trajectory branches are trained using
the predicted trajectories. Both video models start from the same released
DreamDojo-2B 140K checkpoint. Each trajectory branch copies the query, key,
and value projections and the query and key normalization parameters from
the corresponding backbone block. On Bridge, we retain the frozen native
latent action module. On LIBERO, we disable this module and reinitialize
the input layer of the action embedding for the LIBERO action interface.

\begin{table}[tbp]
\centering
\caption{\textbf{ActTrail training settings for main-paper Table~1.}
Updates are counted separately for each stage. Both stages use one
optimizer update per batch, with no gradient accumulation.}
\label{tab:sup-training}
\small
\setlength{\tabcolsep}{2pt}
\begin{tabular}{@{}p{.25\linewidth}p{.31\linewidth}p{.39\linewidth}@{}}
\toprule
Setting & \raggedright Stage~1\newline Trajectory predictor &
\raggedright Stage~2\newline Video generator \tabularnewline
\midrule
Optimizer & \raggedright AdamW & \raggedright Fused AdamW \tabularnewline
Adam $(\beta_1,\beta_2)$ & $(0.9,0.95)$ & $(0.9,0.999)$ \tabularnewline
\raggedright Weight decay / clip norm & $0.01$ / $1$ & $0.1$ / $0.1$ \tabularnewline
\raggedright Learning rate &
\raggedright 200-step linear warmup to $10^{-4}$, then constant &
\raggedright Base $3\times10^{-5}$, 100-step linear warmup,
then linear decay on a 400K-step schedule \tabularnewline
\raggedright Global batch & 2 clips & 2 clips \tabularnewline
Bridge &
\raggedright 1 GPU, 2 clips per GPU, 30K updates &
\raggedright 2 GPUs (DDP), 1 clip per GPU, 30K updates \tabularnewline
LIBERO &
\raggedright 1 GPU, 2 clips per GPU, 130K updates &
\raggedright 1 GPU, 2 clips per GPU, 130K updates \tabularnewline
Precision &
\raggedright BF16 autocast, FP32 parameters and loss &
\raggedright BF16 backbone and fusion autocast, FP32 fusion weights and geometry \tabularnewline
\bottomrule
\end{tabular}
\end{table}

Both optimizers use $\epsilon=10^{-8}$. At Stage~2 update $k$ (counting from one),
the learning rate is $3\times10^{-5}\,r(k-1)$. The multiplier $r(n)$
rises linearly from $r(0)=10^{-6}$ to $r(100)=0.99$, then falls linearly
to $r(400{,}000)=0.4$. Table~\ref{tab:sup-training} gives the training
duration for each dataset.

\noindent\textbf{Video training objective.}
Let $z$ denote a clean video latent and let $\epsilon\sim\mathcal{N}(0,I)$.
We draw one noise level per clip from a logit-normal distribution and apply
the scheduler's shift of 5 to obtain $s$. The rectified-flow interpolation
and target velocity are $z_s=(1-s)z+s\epsilon$ and $u=\epsilon-z$.
We replace the network input at slot~0 with the initial-frame conditioning
latent. After 500 training steps, we increase the standard deviation of the
Gaussian perturbation applied to this condition to 0.03 over the next
1,500 steps. Before computing the loss, we replace the predicted velocity
at slot~0 with its analytical target. This gives zero error at slot~0
while retaining its entries in the denominator of the mean.

Let $e=\hat{u}-u$ and let $e_c$ denote the velocity error in latent
channel $c$. The video objective is
\begin{equation}
\mathcal{L}_{\mathrm{video}}
=\operatorname{mean}(e^2)
+0.1\,\operatorname{mean}\!\left[(e_{c+1}-e_c)^2\right].
\label{eq:sup-video-objective}
\end{equation}
The auxiliary term compares adjacent latent channels at each time and
spatial position, averaging over clips, channel pairs, and positions.

\noindent\textbf{Inference and baseline sampling.}
ActTrail predicts trajectories once from the initial frame and actions.
Each denoising step samples the current DiT features at the same coordinates
to generate all 12 future frames jointly. For Table~1 of the main paper,
all video models use 30K training updates on Bridge and 130K on LIBERO.
Table~\ref{tab:sup-sampling} gives the samplers. For DreamDojo and ActTrail,
let $u_c$ and $u_u$ be the conditional and unconditional velocities.
Their guided velocity is $u_c+g(u_c-u_u)$, equivalent to a conventional
guidance scale of $1+g$ (1 on Bridge and 4 on LIBERO).

\begin{table}[tbp]
\centering
\caption{\textbf{Video sampling settings for main-paper Table~1.}
Step counts follow each method's sampler.}
\label{tab:sup-sampling}
\small
\setlength{\tabcolsep}{2.5pt}
\begin{tabular}{@{}p{.23\linewidth}p{.73\linewidth}@{}}
\toprule
Method & Sampler and guidance \tabularnewline
\midrule
\raggedright DreamDojo, ActTrail &
\raggedright FlowUniPC, 35 joint denoising steps per 12-frame prediction,
shift 5. Bridge uses $g=0$ and LIBERO uses $g=3$. \tabularnewline
IRASim &
\raggedright PNDM, 50 requested steps. The startup phase requires
multiple evaluations, giving 59 denoiser calls per clip. Guidance 1.
\tabularnewline
Vid2World &
\raggedright Autoregressive DDIM, 50 steps per future frame
(600 sampling steps for 12 frames). Action guidance 2.5,
guidance rescale 0.7, $\eta=1$, uniform-trailing timestep spacing,
and history noise level 20. \tabularnewline
Ctrl-World &
\raggedright SVD Euler, 50 steps, guidance 1. Frame-rate conditioning 7
and motion bucket 127. \tabularnewline
\bottomrule
\end{tabular}
\end{table}

\section{Additional Analyses}
\label{sec:supp-additional-analyses}

\subsection{Motion accuracy from predicted trajectories to generated videos}
\label{sec:supp-stage-motion}

ActTrail first predicts where points in the initial frame will move
and then uses these trajectories to guide video generation.
To examine how closely the predicted and generated motion follow the
recorded motion, we compare three trajectories for the same robot
foreground points on the BridgeData V2 test set. We obtain one
trajectory directly from ActTrail's frozen Stage~1 predictor and track
the other two in videos generated by DreamDojo and ActTrail, respectively.
For all three, we use the trajectory tracked in the recorded video as
the reference.

\begin{table}[tbp]
    \centering
    \caption{\textbf{Motion accuracy on BridgeData V2.}
    Each metric uses points and time steps that are valid for the
    reference and all three predictions. Step is measured in pixels
    per interval. Mean cumulative and endpoint errors are measured
    in pixels. Step and Mean cum.\ cover 122 test clips, and End
    covers 121.}
    \label{tab:supp-stage-motion}
    \small
    \setlength{\tabcolsep}{2.5pt}
    \begin{tabular*}{\linewidth}{@{\extracolsep{\fill}}lrrr@{}}
        \toprule
        Trajectory source
        & Step $\downarrow$
        & Mean cum. $\downarrow$
        & End $\downarrow$ \\
        \midrule
        DreamDojo video
        & 13.69 & 45.03 & 68.14 \\
        ActTrail Stage~1
        & \textbf{8.60} & 30.06 & 44.31 \\
        ActTrail Stage~2 video
        & 10.65 & \textbf{25.37} & \textbf{37.27} \\
        \bottomrule
    \end{tabular*}
\end{table}

ActTrail's final videos have lower errors than DreamDojo on all three
metrics (Table~\ref{tab:supp-stage-motion}). Within ActTrail, Stage~1
predicts frame-to-frame displacements more accurately, yet points in
the final videos lie closer to their recorded positions on average.
Small displacement errors can accumulate when successive steps err
in similar directions. For points valid in all 13 frames (120 clips),
the mean cosine similarity between consecutive stepwise error vectors
was 0.234 for Stage~1 and 0.079 for the final videos.
For each point, we divide its endpoint error by the sum of its
12 stepwise error magnitudes. This fraction averages 38.5\% for
Stage~1 and 26.7\% for the final videos, indicating that the larger
stepwise errors in the videos cancel to a greater extent over time.
On this continuously tracked subset, the final videos retain their
advantage in mean cumulative error. Their mean endpoint error is
also lower, although the paired 95\% confidence interval for this
difference includes zero ($-6.30$~px, $[-13.76,\,0.16]$).

The improvement in position accuracy is clearest along the direction
of recorded motion. At each step, we use the direction from a point's
initial position to its current recorded position to measure how far
the prediction leads or falls short and how far it deviates sideways.
At step~12, averaging overshoots and undershoots with opposite signs,
Stage~1 predictions fall short by 24.32~px on average, compared with
11.37~px for the final videos. The final videos also have smaller
errors along this direction when we take the absolute value for each
point before averaging. Lateral errors show no clear improvement.
Thus, the final videos more closely reproduce how far the robot points
move along the recorded direction, even though Stage~1 predicts their
individual frame-to-frame displacements more accurately.

\subsection{Using predicted trajectories to guide video features}
\label{sec:supp-fusion}

A predicted trajectory gives both the future positions of a scene point
and the correspondence between those positions. We compare three ways
of using this information on BridgeData V2. All three models use the
same frozen Stage~1 predictor and are trained for 30,000 video-model
updates.

Consider a point on the gripper. B-Mod encodes its coordinates and
displacements as modulation parameters for video features. Feature
Warping transports the point's initial-frame features to its predicted
positions and combines them with the video features there. ActTrail
reads the current video features at each predicted position, updates
them through attention along the path of the same point, and writes
each update back to its corresponding position. The four updates that
share a latent time slot are combined after this spatial write-back.

\begin{figure*}[t]
    \centering
    \includegraphics[width=\textwidth]{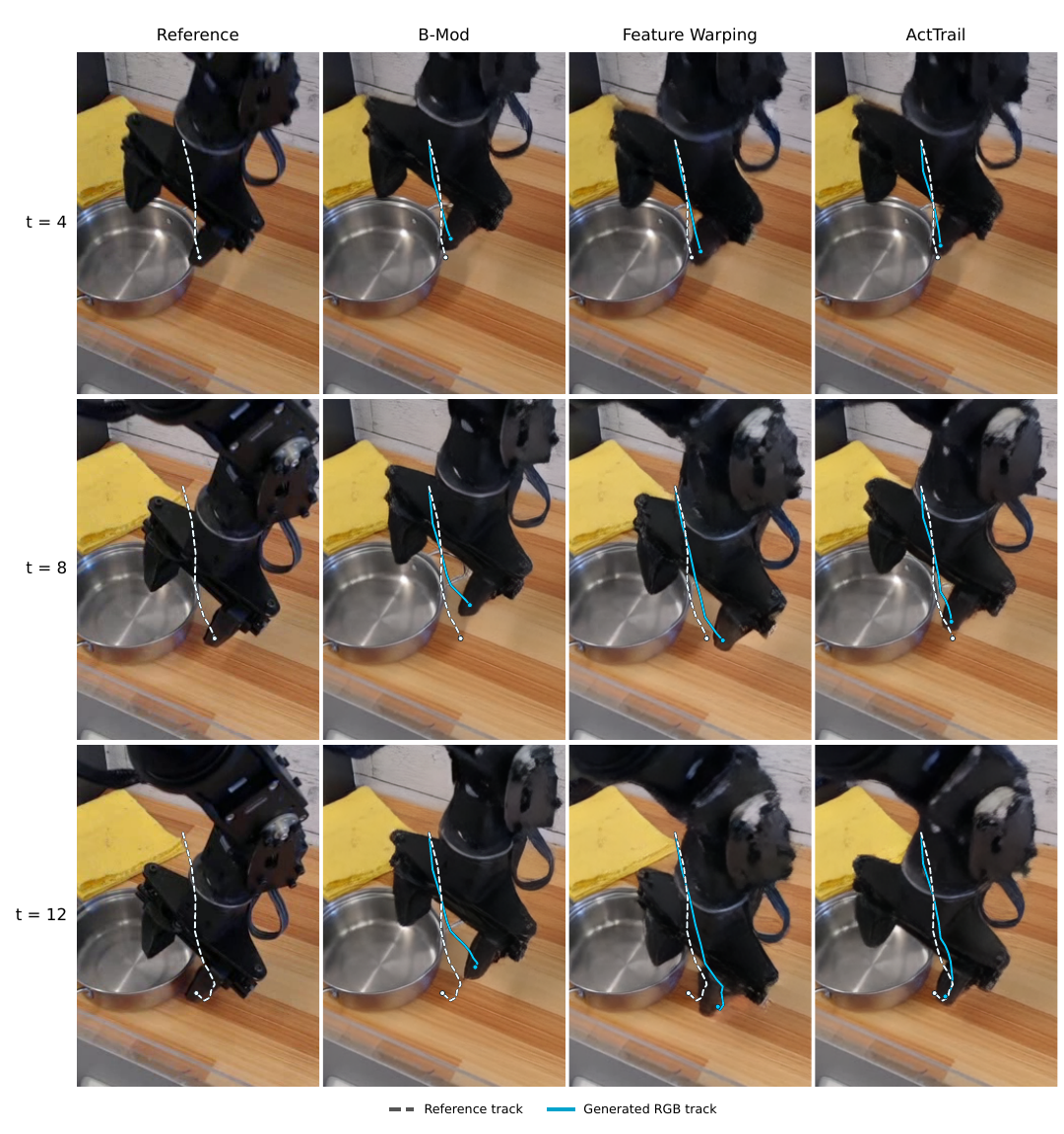}
    \caption{\textbf{Gripper motion under three trajectory fusion designs
    on BridgeData V2.} Each row shows the same test clip at the indicated
    step. The dashed line traces a gripper point in the recorded video.
    In the three generated-video columns, the solid cyan line traces the
    corresponding point from the initial frame to the displayed step.}
    \label{fig:supp-fusion-tracks}
\end{figure*}

\begin{table}[tbp]
    \centering
    \caption{\textbf{Trajectory fusion on BridgeData V2.}
    Step is measured in pixels per interval and End in pixels at step~12.
    Dyn.~L1 is RGB error within moving-region interiors on the
    0--255 scale. Lower is better for all three metrics.}
    \label{tab:supp-fusion}
    \small
    \setlength{\tabcolsep}{3pt}
    \begin{tabular*}{\linewidth}{@{\extracolsep{\fill}}lrrr@{}}
        \toprule
        Fusion design & Step $\downarrow$ & End $\downarrow$
                      & Dyn.~L1 $\downarrow$ \\
        \midrule
        B-Mod           & 12.42 & 42.69 & 31.26 \\
        Feature Warping & 10.93 & 36.25 & 30.69 \\
        ActTrail        & \textbf{10.87} & \textbf{35.16} & \textbf{29.90} \\
        \bottomrule
    \end{tabular*}
\end{table}

Using predicted positions to place visual features yields lower mean
motion and appearance errors than coordinate modulation
(Table~\ref{tab:supp-fusion}). Feature Warping and ActTrail have
similar stepwise errors. ActTrail achieves the lowest mean endpoint
error and the lowest mean RGB error in moving regions. The final row
of Figure~\ref{fig:supp-fusion-tracks} illustrates the spatial
difference. In this clip, the recorded gripper path bends to the left
near its endpoint, and the ActTrail video follows this bend more
closely than the other two predictions.

\subsection{Combining action steps within a shared temporal slot}
\label{sec:supp-within-slot}

As the gripper moves through four successive positions, ActTrail writes
a feature update at each predicted position. These four action steps
share one temporal slot in the video model, so their updates must
ultimately be combined. On BridgeData V2, we compare Full and PShared
to assess whether accounting for a step's position within its slot
improves the motion in the generated video.

Both designs retain all 13 trajectory nodes, including the initial
node. They perform attention along each point's trajectory and write
features back at all 12 future positions. PShared applies the same
affine transformation to the features read at all four steps within
a slot and averages the four updates after write-back. Full instead
learns a distinct affine transformation and a vector of channel-wise
aggregation weights for each of the four steps. Each DiT block learns
its own four sets of parameters and reuses them across its three future
slots. For example, steps~1, 5, and~9 use the same parameters within a
block because each is the first step in its slot.

\begin{table}[tbp]
    \centering
    \caption{\textbf{Within-slot feature processing on BridgeData V2.}
    Step measures frame-to-frame displacement error in pixels per
    interval, and End measures endpoint error at step~12 in pixels.
    These metrics use tracks valid in the reference and both predictions,
    covering 122 and 120 test clips, respectively. Dyn.~L1 measures
    RGB error within moving-region interiors on the 0--255 scale
    across all 128 clips.}
    \label{tab:supp-within-slot}
    \small
    \setlength{\tabcolsep}{3pt}
    \begin{tabular*}{\linewidth}{@{\extracolsep{\fill}}lrrr@{}}
        \toprule
        Within-slot design & Step $\downarrow$ & End $\downarrow$
                           & Dyn.~L1 $\downarrow$ \\
        \midrule
        PShared       & 11.07          & 38.92          & \textbf{29.62} \\
        ActTrail Full & \textbf{10.84} & \textbf{37.05} & 29.90          \\
        \bottomrule
    \end{tabular*}
\end{table}

Although both designs write features back at the same predicted
positions, Full more accurately reproduces the recorded frame-to-frame
displacements and reduces the mean endpoint error by 1.87~pixels
relative to PShared (Table~\ref{tab:supp-within-slot}). PShared yields
a slightly lower RGB error in moving regions. The gain from Full is
therefore clearest in motion accuracy: with per-step spatial write-back
in place, the results support learning how to transform and combine
features according to each step's position within the shared slot.

\section{Gripper Endpoints under Paired Actions}
\label{sec:supp_libero_endpoints}

\begin{figure*}[t]
    \centering
    \includegraphics[width=\textwidth]
        {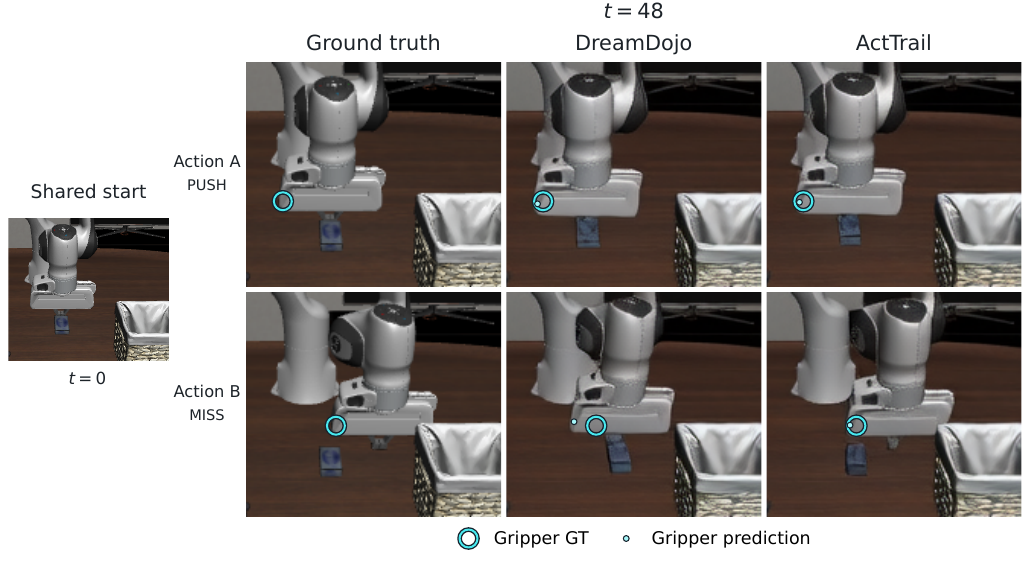}
    \caption{\textbf{Gripper endpoints under paired actions on LIBERO.}
    The left panel shows the shared initial frame. The two rows show
    the simulator outcome and both model predictions at step~48 for
    the \textsc{Push} and \textsc{Miss} branches. Open circles mark
    the reference endpoint of the same gripper point in each branch.
    Filled dots mark its location in the generated frames.}
    \label{fig:supp_libero_gripper_endpoints}
\end{figure*}

To illustrate how the choice of action affects the final gripper
position, we examine one example from the LIBERO paired-action
evaluation in Table~2 of the main paper. Two action sequences start
from the same simulator state. One directs the gripper toward the
box (\textsc{Push}), while the other takes it past the box
(\textsc{Miss}). For each branch, DreamDojo and ActTrail receive the
same initial frame and action sequence. Each model predicts 48 steps
in four 12-step segments, using its own final frame from the preceding
segment to start the next segment within the same branch.
Figure~\ref{fig:supp_libero_gripper_endpoints} shows the predictions
and simulator outcomes at step~48.

For \textsc{Push}, both models place the gripper close to its reference
endpoint. Across the jointly valid gripper points in this branch, the
endpoint errors are 3.43 pixels for DreamDojo and 2.01 pixels for
ActTrail. The difference is clearer for \textsc{Miss}. From the shared
initial frame to step~48, the marked gripper point moves 34.73 pixels
to the right in the simulator. DreamDojo predicts a rightward
displacement of 20.15 pixels, while ActTrail predicts 30.53 pixels.
Both models capture the rightward motion, but DreamDojo underestimates
its extent. These horizontal displacements refer to the single marked
point in the original image coordinates. Averaged over all jointly
valid gripper points in the \textsc{Miss} branch, the 2D endpoint
error is 14.99 pixels for DreamDojo and 3.62 pixels for ActTrail.

\section{Motion Prediction in Real-Robot Manipulation}
\label{sec:supp_real_robot_outcomes}

Using a predicted video to assess the likely outcome of a robot
action requires seeing how the end effector moves relative to the
object being manipulated. We evaluate 36-step video predictions
on three real-robot tasks. In grasping trials, we randomized the
initial position of the centrifuge tube. In placement trials, we
randomized the target location for the tube inside the centrifuge.
In pipette-tip attachment trials, we randomized the position of the
tip to be attached. For each trial, DreamDojo and ActTrail receive
the same recorded initial frame and executed action sequence, and
we compare both predictions with the recording of that trial.
As reported in Table~3 of the main paper, ActTrail has lower
stepwise displacement and endpoint errors than DreamDojo on all
three tasks. The largest difference in endpoint error occurs in
pipette-tip attachment. Across the 16 test trials, mean endpoint
error is 20.573~pixels for DreamDojo and 3.165~pixels for ActTrail.

Video~S1 shows one trial in which the tip remained in the rack.
The recording and the two predictions are shown side by side,
synchronized by action step, with a $3\times$ enlargement of the
same fixed region below each view. During steps 24--36, the
recorded pipette lifts away while the tip stays in its original
rack slot, leaving a widening gap between them. ActTrail also
shows this gap opening as the pipette rises. In the DreamDojo
prediction, the pipette end continues to overlap the tip opening
in the image, making it harder to see when the two separate.

%% file: preamble.tex
\usepackage{multirow}



%% file: fig/teaser.tex
\begin{strip}
    \centering
    \includegraphics[width=\textwidth]{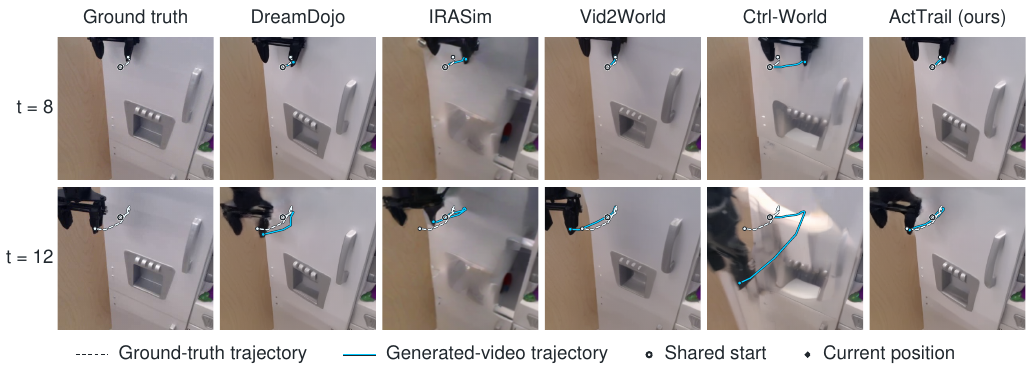}
    \captionsetup{hypcap=false}
    \captionof{figure}{\textbf{Action-conditioned video prediction on
    BridgeData V2.}
    All methods receive the same initial frame and action sequence.
    Rows show steps 8 and 12.
    Overlaid trajectories track a gripper point in the recorded
    and generated videos; markers indicate the shared starting
    point and current positions.}
    \label{fig:acttrail_teaser}
\end{strip}

%% file: sec/0_abstract.tex
\begin{abstract}
Video world models enable robots to anticipate action consequences,
yet visually realistic predictions can still depict motion that
differs from actual execution.
We introduce ActTrail, which predicts cumulative 2D displacements
for a dense set of initial-frame points from an initial RGB image
and an action sequence.
This translates low-dimensional robot controls into explicit
multi-step motion guidance for video generation.
Point tracks and adjacent-frame optical flow extracted from
training videos provide complementary supervision for cumulative
displacement and stepwise motion, respectively.
A trajectory fusion module aggregates visual features along
predicted paths, writes per-step updates back to their spatial
locations, and merges them within the latent time slots of a
video diffusion transformer, preserving action-step motion
guidance under temporal compression.
Experiments on BridgeData V2 and LIBERO demonstrate improved
motion fidelity and visual quality in dynamic regions, including
a 42.4\% reduction in mean cumulative displacement error on
BridgeData V2 relative to DreamDojo.
Across 52 real-robot trials, human success/failure judgments
based on ActTrail's predicted videos agree with recorded
execution outcomes in 94.2\% of cases, compared with 82.7\%
for DreamDojo.
\end{abstract}

%% file: sec/1_intro.tex
\section{Introduction}
\label{sec:intro}
Action-conditioned video world models predict future scenes
from a robot's current observation and a sequence of candidate
actions~\cite{unisim,Zhu_2025_ICCV,vid2world}.
These predictions support the comparison of action plans before
execution~\cite{Bar_2025_CVPR,pmlr-v306-gao26q} and the closed-loop
evaluation of vision-language-action (VLA)
policies~\cite{guo2026ctrlworld}.
Such applications require generated videos to accurately depict
changes in the robot and the objects it manipulates throughout execution.
Yet even visually realistic predictions can show motion whose
direction, distance, or final position deviates from actual
execution~\cite{worldsync}.
For two action sequences that would produce clearly different
outcomes from the same initial state, a model may nevertheless
generate similar motion.
The central challenge is therefore to accurately predict how
the robot and objects move across successive frames under the
given actions while maintaining visual quality in moving regions,
so that the videos can reliably support plan comparison and
task-outcome assessment.

Recent action-conditioned video world models improve control
over generated videos through frame-level or grouped action
conditioning~\cite{Zhu_2025_ICCV,pmlr-v306-gao26q}.
These designs strengthen the temporal alignment between actions
and video frames, but leave the video generator to infer how
the robot and objects move in the image.
Other approaches explicitly represent action-induced motion
using scene flow or point trajectories to provide spatial
guidance for video generation, but require depth information
or rely on camera calibration and simulation for forward
prediction~\cite{guo2026flowdreamer,li2026hydra0}.

To this end, we introduce ActTrail, which uses only an initial
RGB image and a given action sequence to predict dense 2D point
trajectories for future video generation.
This makes scene-point motion explicit without requiring
additional geometric inputs or simulation at inference.
Point tracks and adjacent-frame optical flow extracted from
training videos supervise cumulative displacement and stepwise
motion, respectively, to improve both multi-step position
accuracy and the prediction of motion between adjacent frames.
Our fusion module adapts trajectory-based feature
aggregation~\cite{xiao2025trajectoryattention} to the video
model's temporal compression.
It preserves action-step positions and cross-step
correspondences before merging trajectory-guided updates
into shared latent time slots.

We evaluate motion prediction accuracy and visual quality in
moving regions on BridgeData V2~\cite{walke2023bridgedata} and
LIBERO~\cite{liu2023libero}.
Paired-action tests assess responses to alternative actions
from the same initial state, while real-robot experiments
evaluate agreement with recorded motion, appearance,
and task outcomes.

The main contributions of our work are as follows:
\begin{itemize}
    \item We introduce ActTrail, an action-conditioned video
    world model that predicts dense 2D point trajectories to
    guide future video generation.

    \item We combine point-track and adjacent-frame optical-flow
    supervision to constrain cumulative displacement and
    stepwise motion, respectively.
    Our fusion scheme retains per-action-step positions and
    correspondences in a temporally compressed video DiT.

    \item Across both benchmarks, ActTrail achieves the lowest
    mean cumulative displacement and moving-region interior
    RGB $L_1$ errors among the compared methods.
    Paired-action results further show more accurate predictions
    of appearance and gripper-motion differences between
    action branches.
    Across the real-robot tasks, human success/failure judgments
    based on ActTrail's videos agree more often with recorded
    execution outcomes than those based on DreamDojo.
\end{itemize}

%% file: sec/2_related_work.tex
\section{Related Work}
\label{sec:related_work}

\subsection{Action-Conditioned World Models for Robotics}
\label{sec:related_action_world_models}

Action-conditioned world models predict environmental changes
caused by control commands, supporting robotic planning and
policy evaluation.
Diverse interaction data and large-scale video pretraining
improve model capabilities~\cite{unisim,pmlr-v306-gao26q},
while latent actions learned from videos facilitate adaptation
across action spaces~\cite{adaworld}.
Frame-level or causal action conditioning strengthens temporal
alignment between actions and predicted
frames~\cite{Zhu_2025_ICCV,vid2world}, and multi-view observations
with historical context support sustained interaction and
policy evaluation~\cite{guo2026ctrlworld}.
To address discrepancies between generated videos and given
actions, WorldEcho extends evaluation to off-expert actions,
while WorldSync improves action responsiveness through more
diverse action coverage, dynamics supervision, and
intervention-effect alignment~\cite{worldsync}.

Explicit motion approaches differ in their prediction targets
and in how motion conditions are obtained.
Three-dimensional world models directly predict scene-point
displacements or jointly model actions and scene motion in
a shared space~\cite{pointworld,pointwam,unitas}.
For future video generation, motion guidance can come from
scene flow predicted from RGB-D observations and
actions~\cite{guo2026flowdreamer}, projections of simulated
robot motion or renderings derived from known robot
models~\cite{li2026hydra0,robotfactored}, or 3D point trajectories
describing manipulation contacts~\cite{contactflow}.
These approaches provide explicit spatial conditions through
scene geometry, robot motion, or contact information.

Optical flow and visual tracks also connect video prediction
with action generation or selection~\cite{flowwam,tract}.
ActTrail focuses on predicting future scene-point motion
and RGB video from an initial image and a specified
robot action sequence.

\subsection{Trajectory-Guided Video Generation}
\label{sec:related_video_motion_control}

Motion control in video generation constrains visual changes
through motion vectors, camera poses, or object
trajectories~\cite{videocomposer,motionctrl}.
Recent work supports trajectory conditions of varying densities
and granularities, enabling control over individual objects,
multiple objects, and global scene
motion~\cite{motionprompting,magicmotion,flextraj}.

Fusion strategies encode trajectories as spatiotemporal
conditions for video DiT blocks~\cite{tora} or process
conditioning sequences jointly with video
representations~\cite{flextraj}.
Cross-frame correspondences along trajectories also support
visual feature propagation and interaction:
Motion-I2V~\cite{motioni2v} predicts pixel trajectories from
an image and text, then uses them to propagate reference-image
features; Wan-Move~\cite{wanmove} propagates first-frame features
along trajectories to future latent positions; and Trajectory
Attention~\cite{xiao2025trajectoryattention} performs attention
along motion paths.

Our fusion scheme adapts trajectory-guided feature
interaction to temporal compression, retaining distinct
predicted positions for action steps that share a latent
time slot.

%% file: sec/3_method.tex
\section{Method}
\label{sec:method}

\input{figures/figure2_latex}

\subsection{Task Formulation and Overview}
\label{sec:task_overview}

\textbf{Problem formulation.}
Given an initial RGB image
$I_0\in\mathbb{R}^{H\times W\times3}$
and a robot action sequence $a_{1:T}$, we aim to predict
future video frames $\hat I_{1:T}$ that accurately depict
the motion of the robot and the objects it interacts with.
Each action $a_t$ corresponds to the interval between
sampled frames $I_{t-1}$ and $I_t$.
We set $T=12$ for each prediction window.

\textbf{Overview.}
As illustrated in \cref{fig:overview}, ActTrail connects
robot actions to future video generation through predicted
point trajectories.
Given the initial frame and action sequence, the trajectory
predictor $f_\theta$ estimates cumulative 2D displacements
$\hat P_{1:T}$ of dense scene points relative to their
initial positions.
The video generator $p_\phi$ builds on
DreamDojo~\cite{pmlr-v306-gao26q}, retaining its original
image and action conditioning while using the predicted
trajectories to guide generation.
The overall process is
\begin{equation}
\begin{aligned}
\hat P_{1:T} &= f_\theta(I_0,a_{1:T}),\\
\hat I_{1:T} &\sim p_\phi\!\left(
    \cdot\mid I_0,a_{1:T},\hat P_{1:T}
\right).
\end{aligned}
\label{eq:ActTrail_overview}
\end{equation}

\subsection{Action-Conditioned Trajectory Prediction}
\label{sec:trajectory_prediction}

\textbf{Trajectory representation.}
We define a regular grid of points $\{q_i\}_{i=1}^{N}$ in
the initial frame and preserve their identities across time.
The predictor directly estimates each point's cumulative
2D displacement $\hat p_{t,i}$ from its initial position.
The corresponding position and stepwise displacement are
\begin{equation}
\hat q_{t,i}=q_i+\hat p_{t,i},
\qquad
\hat v_{t,i}=\hat p_{t,i}-\hat p_{t-1,i},
\label{eq:trajectory_representation}
\end{equation}
where $\hat p_{0,i}=0$.
Positions and displacements are expressed in coordinates
of the $60\times80$ trajectory grid, with cumulative
displacements for all $T$ steps predicted in a single
forward pass.

\textbf{Prediction network.}
As shown in \cref{fig:overview}(a), a frozen DINOv2-Large
encoder~\cite{oquab2024dinov2} extracts initial-frame features,
which are aligned to the trajectory grid through bilinear
sampling and a learnable projection.
An MLP and a causal Transformer encode the actions so that
the condition at step $t$ depends only on $a_{1:t}$.
Channel-wise scales and biases derived from these conditions
modulate features within a spatial U-Net shared across steps.
A causal temporal residual module then combines current and preceding features for each initial-frame point. A shared \(1\times1\) convolutional head predicts the cumulative displacements.

\textbf{Trajectory and flow supervision.}
Point tracks and adjacent-frame optical flow constrain
cumulative displacement and stepwise motion, respectively.
We sample teacher trajectories $q^\star_{t,i}$ for the
initial grid points from AllTracker
predictions~\cite{Harley_2025_ICCV}, defining the cumulative
targets as $p^\star_{t,i}=q^\star_{t,i}-q_i$, with
$q^\star_{0,i}=q_i$.
For each interval, we map the preceding teacher position
$q^\star_{t-1,i}$ to image coordinates, sample the forward
optical flow estimated by DPFlow~\cite{Morimitsu_2025_CVPR},
and convert the resulting vector to grid displacement units
to obtain $v^\star_{t,i}$.
The two targets therefore refer to the same initial-frame
points; the flow targets are not temporal differences
of the teacher tracks.

Let $\Omega$ contain the jointly valid sample, timestep,
and point indices for both targets in a training batch.
Using the Charbonnier penalty
$\rho(r)=\sqrt{\lVert r\rVert_2^2+10^{-6}}$,
we define the losses and combined objective as
\begin{equation}
\begin{aligned}
L_P
&=\frac{1}{|\Omega|}
\sum_{(b,t,i)\in\Omega}
\rho\!\left(\hat p_{b,t,i}-p^\star_{b,t,i}\right),\\
L_{\mathrm{flow}}
&=\frac{1}{|\Omega|}
\sum_{(b,t,i)\in\Omega}
\rho\!\left(\hat v_{b,t,i}-v^\star_{b,t,i}\right),\\
L_{\mathrm{motion}}
&=L_{\mathrm{flow}}+\lambda L_P,
\end{aligned}
\label{eq:motion_objective}
\end{equation}
where $b$ indexes samples and $\lambda=0.1$ is fixed.

\subsection{Trajectory-Guided Video Generation}
\label{sec:trajectory_guided_generation}

As shown in \cref{fig:overview}(b,c), we introduce a trajectory
branch into the video DiT to update video features using
predicted positions and correspondences across action steps.

\begin{figure}[t]
    \centering
    \includegraphics[width=\columnwidth]
        {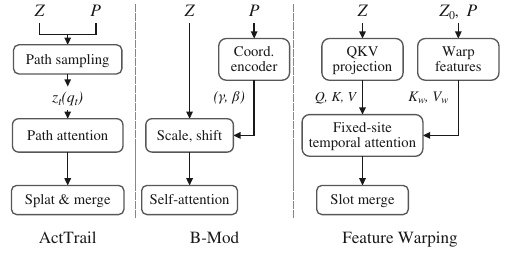}
    \caption{\textbf{Trajectory fusion in Stage~2.}
    ActTrail attends along predicted trajectories,
    B-Mod modulates video features, and Feature Warping,
    adapted from Motion-I2V~\cite{motioni2v},
    transports first-frame features to predicted locations.
    $Z$, $Z_0$, and $P$ denote video features,
    first-frame features, and predicted trajectories,
    respectively.}
    \label{fig:stage2_fusion}
\end{figure}

\textbf{Trajectory--video alignment.}
Predicted trajectories specify spatial positions at each
action step, whereas each future latent time slot in the
video backbone represents four consecutive steps.
We map the initial frame and future steps to latent time
slots as
\begin{equation}
g(t)=
\begin{cases}
0, & t=0,\\
\lceil t/4\rceil, & 1\leq t\leq T.
\end{cases}
\label{eq:trajectory_slot_mapping}
\end{equation}
We use the centers of the first-frame video tokens as source
points indexed by $j$.
We express each source position in trajectory-grid coordinates
and bilinearly sample its cumulative displacements on the
$60\times80$ grid.
We then add these displacements to the source position and
map the resulting positions to the $30\times40$ video feature
grid, obtaining $\hat{x}_{t,j}$.
Nodes assigned to the same slot access a shared feature map
at their respective predicted positions.

\textbf{Trajectory-guided feature interaction.}
Building on Trajectory Attention~\cite{xiao2025trajectoryattention},
which associates features along motion paths, we adapt this
operation to action-step trajectories and temporally
compressed video features.
Let $F_k^{(\ell)}$ denote the feature map in DiT block $\ell$
after normalization and conditional modulation but before
self-attention.
Features sampled along each source point's predicted
trajectory form a sequence for interaction:
\begin{equation}
\begin{aligned}
z_{t,j}^{(\ell)}
&=\operatorname{Bilinear}
\!\left(F_{g(t)}^{(\ell)},\hat x_{t,j}\right),\\
\widetilde Z_j^{(\ell)}
&=\mathcal A^{(\ell)}
\!\left(Z_j^{(\ell)},\hat X_j\right),
\end{aligned}
\label{eq:trajectory_feature_interaction}
\end{equation}
where $Z_j^{(\ell)}=\{z_{t,j}^{(\ell)}\}_{t=0}^{T}$ and
$\hat X_j=\{\hat x_{t,j}\}_{t=0}^{T}$ contain features and
positions for all $T+1$ nodes, including the initial frame.

The operator $\mathcal A^{(\ell)}$ first applies a
phase-dependent channel-wise affine transformation to the
sampled features, then performs attention independently
along each source point's trajectory.
A phase denotes a step's position within its four-step group.
The affine parameters are shared across future slots within
each block, with separate parameters for the initial frame.
Spatiotemporal rotary position embeddings encode
$\hat x_{t,j}$ and the temporal coordinate $t/4$ to
distinguish nodes within a shared slot.
A slot-level prefix mask allows a node at step $t$ to attend
only to nodes at steps $s$ satisfying $g(s)\leq g(t)$;
nodes within the same slot can attend to one another.
The geometric mask $m_{t,j}$ excludes non-finite or
out-of-bounds coordinates and does not estimate visibility.
Invalid nodes contribute neither to attention aggregation
nor to spatial write-back.

\textbf{Spatial write-back and temporal aggregation.}
Bilinear forward splatting writes the features updated by trajectory attention back to the video grid at each step’s predicted positions.
Normalization is performed separately at each action step:
the accumulated features at each destination are divided
by the total splatting weight received there.
These stepwise updates are then combined within their
corresponding latent time slots:
\begin{equation}
\begin{aligned}
R_t^{(\ell)}
&=\operatorname{SplatNorm}
\!\left(
\{\widetilde z_{t,j}^{(\ell)},\hat x_{t,j},m_{t,j}\}_{j}
\right),\\
\Delta H_k^{(\ell)}
&=W_{\mathrm{out}}^{(\ell)}
\!\left(
\sum_{t\in\mathcal G_k}
w_{r(t)}^{(\ell)}\odot R_t^{(\ell)}
\right),\\
H_k^{(\ell)}
&\leftarrow H_k^{(\ell)}+\Delta H_k^{(\ell)}.
\end{aligned}
\label{eq:trajectory_writeback_aggregation}
\end{equation}
Here, $\operatorname{SplatNorm}$ accumulates contributions
only from valid nodes and assigns zero to locations
receiving no contribution.
The set $\mathcal G_k=\{t>0:g(t)=k\}$ contains the action
steps assigned to future slot $k$, and
$r(t)=1+((t-1)\bmod4)$ identifies the within-slot phase.
The weights $w_{r(t)}^{(\ell)}$ are learned per phase and
channel, shared across future slots within each block,
and applied directly without weight normalization.
The operator $\odot$ denotes channel-wise multiplication.

The output projection $W_{\mathrm{out}}^{(\ell)}$ is
zero-initialized, and $H_k^{(\ell)}$ denotes the features
after the backbone's original gated self-attention
residual update.
The trajectory residual is added before cross-attention,
after which the backbone proceeds with its original
cross-attention and feed-forward network.
Initial-frame nodes participate in trajectory interaction,
while the additional residual for their time slot remains
zero.

\subsection{Training and Inference}
\label{sec:training_inference}

\textbf{Training strategy.}
ActTrail is trained in two stages.
First, we freeze the DINOv2 encoder and optimize the
trajectory predictor using the trajectory and optical-flow
supervision described in \cref{sec:trajectory_prediction}.
The predictor and VAE are then kept frozen while the
video DiT and trajectory branch are trained, conditioned
on trajectories predicted from the initial frame and actions.

\textbf{Video training objective.}
Following the DreamDojo training implementation~\cite{pmlr-v306-gao26q},
we train the video model using rectified-flow velocity regression with an auxiliary loss weighted by 0.1. This auxiliary loss penalizes differences in velocity error between adjacent latent channels.
The initial-frame conditioning slot is masked out of
the prediction error, while the backbone's noise schedule
and conditioning perturbation are retained.

\textbf{Inference.}
Given an initial image $I_0$ and 12 future actions,
the trajectory predictor estimates cumulative
displacements for all future steps in a single forward
pass; these predictions remain fixed throughout denoising.
Conditioned on the initial frame, actions, and predicted
trajectories, the video DiT starts from Gaussian noise.
At each network evaluation, the trajectory branch reads
current video features at the fixed predicted positions
and computes feature updates.
The frozen VAE decodes the generated latents into a
13-frame window, whose last 12 frames constitute the
future predictions.
Inference requires no target trajectories, optical flow,
or future ground-truth frames.

%% file: figures/figure2_latex.tex
\begin{figure*}[t]
    \centering
    \includegraphics[width=\textwidth]{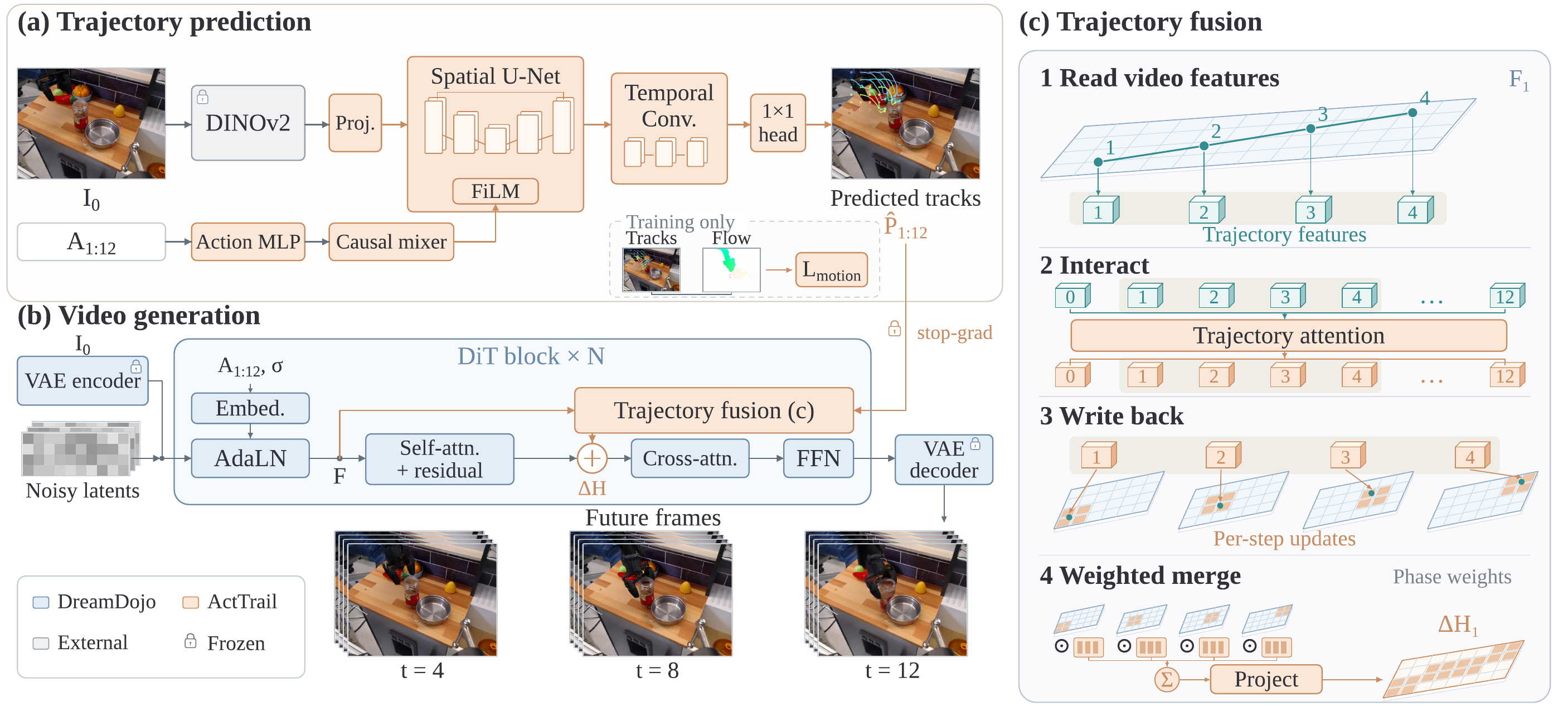}
    \caption{\textbf{Overview of ActTrail.}
    (a) The initial RGB image and actions condition dense
    cumulative 2D displacement prediction, visualized as
    selected point trajectories. Point tracks and adjacent-frame
    optical flow provide training supervision.
    (b) A trajectory branch guides DreamDojo's video DiT,
    retaining its image and action conditioning.
    (c) Features sampled along each point's predicted path
    interact through trajectory attention and are written
    back at individual steps before merging into latent
    time slots. The attention sequence contains all 13 nodes
    with slot-prefix masking; each future slot covers
    four action steps.}
    \label{fig:overview}
\end{figure*}

%% file: sec/4_experiments.tex
\section{Experiments}
\label{sec:experiments}

\subsection{Experimental setup}
\label{sec:experimental_setup}

\noindent\textbf{Data and tasks.}
We evaluate motion fidelity and visual quality on
BridgeData V2~\cite{walke2023bridgedata} and
LIBERO~\cite{liu2023libero}, responses to action changes in
LIBERO, and motion and task-outcome prediction on real-robot data.
For BridgeData V2, we use the IRASim release~\cite{Zhu_2025_ICCV},
with 41,142 training clips from 2,000 trajectories,
64 development clips, and 128 test clips.
Our LIBERO data span 40 tasks and are split by episode
into 1,600 training, 200 development, and 200 test episodes.
We sample four 13-frame clips from each test episode,
yielding 800 evaluation clips.
The standard prediction horizon is 12 steps on both benchmarks.
A separate paired-action evaluation in LIBERO uses 70 initial
states, each with two action branches, to assess 48-step predictions.
Real-robot experiments cover tube grasping, tube placement,
and pipette-tip attachment, with 60/20, 50/16, and 50/16
training/test sequences, respectively.
All three tasks use a 36-step prediction horizon.

\noindent\textbf{Compared methods and training.}
Baselines include DreamDojo~\cite{pmlr-v306-gao26q},
IRASim~\cite{Zhu_2025_ICCV},
Vid2World~\cite{vid2world},
and Ctrl-World~\cite{guo2026ctrlworld}
on both benchmarks, and DreamDojo on real-robot data.
For each test case, all methods receive the same initial image
and action sequence and use the same prediction horizon.
Video models are trained for 30,000 updates on BridgeData V2
and 130,000 updates on LIBERO, and fine-tuned for 5,000 updates
per real-robot task.
ActTrail's trajectory predictor is pretrained separately
and remains frozen during video-model training.

\noindent\textbf{Evaluation metrics.}
We assess motion using frame-to-frame displacement (Step),
mean cumulative displacement (Mean cum.), and endpoint (End)
errors on tracks jointly valid in the reference and all
compared predictions.
PSNR, SSIM, and LPIPS evaluate full-frame visual quality,
while RGB $L_1$ captures reconstruction errors in moving-region
interiors (Dyn.~$L_1$) and inner boundaries (Boundary~$L_1$).
Motion errors are expressed in pixels, or pixels per interval
for Step, with corresponding grid units for Stage~1.
PSNR uses dB, and RGB $L_1$ uses the 0--255 scale.

For paired actions, Acc@2 measures how often a prediction
is closer to its own reference than to the alternative
under full-frame RGB MSE, with half credit for ties.
RGB and Gripper response errors compare predicted and reference
differences between action branches.
Within each branch, Gripper ADE and Gripper FDE quantify
average and final gripper position errors, respectively;
LPIPS-Alex measures perceptual differences between predicted
and reference images.
Paired-action results are weighted by task and action family.

On real-robot data, motion evaluation uses foreground tracks.
Outcome accuracy (Acc.) is the percentage of human
success/failure judgments of predicted videos that match
the recorded execution outcomes.

\subsection{Motion fidelity and visual quality}
\label{sec:bridge_results}

\input{sec/table1_bridge}

We compare 12-step video predictions on BridgeData V2
and LIBERO (\cref{tab:bridge_main}).

\begin{figure}[t]
    \centering
    \includegraphics[width=\columnwidth]{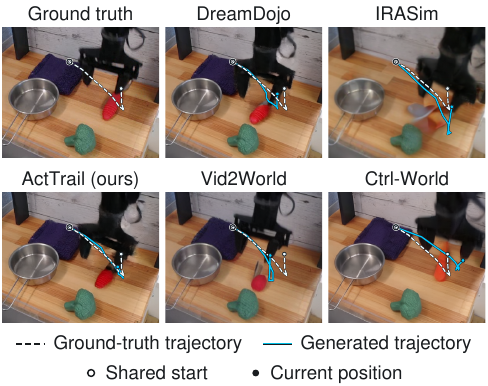}
    \caption{\textbf{Robot motion prediction on BridgeData V2.}
    All panels show step~12 under the same initial-frame
    and action conditions.
    Dashed and solid lines show the reference and generated-video
    gripper trajectories over steps~0--12, respectively.}
    \label{fig:bridge_qualitative}
\end{figure}

\noindent\textbf{Robot motion accuracy.}
On BridgeData V2, ActTrail achieves lower mean cumulative
displacement and endpoint errors than the compared methods.
The LIBERO results also show gains in frame-to-frame and mean
cumulative displacement accuracy.
Qualitative examples illustrate this agreement with recorded
gripper motion across prediction steps
(\cref{fig:acttrail_teaser}) and in the final position
(\cref{fig:bridge_qualitative}).

\begin{figure}[t]
    \centering
    \includegraphics[width=\columnwidth]
        {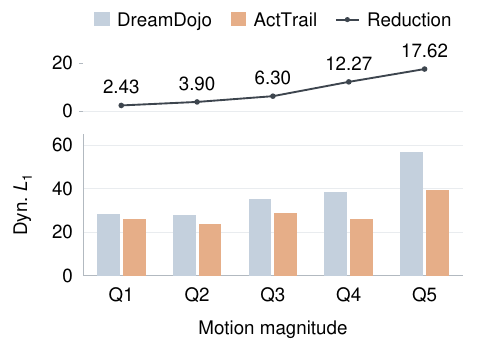}
    \caption{\textbf{Reconstruction error by motion magnitude
    on BridgeData V2.}
    Bars show mean Dyn.~$L_1$ errors, and the upper curve
    shows the absolute reduction (DreamDojo minus ActTrail).
    Q1--Q5 are ordered by increasing reference-motion magnitude,
    with bin thresholds fixed from the training set. Scores are averaged over clips.}
    \label{fig:bridge_motion_bins}
\end{figure}

\noindent\textbf{Visual quality in dynamic regions.}
ActTrail achieves the lowest mean RGB $L_1$ errors in
moving-region interiors and inner boundaries on both
datasets, together with the best full-frame PSNR, SSIM,
and LPIPS scores.
The motion-binned analysis further shows that the absolute
reduction in dynamic-interior RGB $L_1$ relative to DreamDojo
increases with reference-motion magnitude
(\cref{fig:bridge_motion_bins}).

\subsection{Response to action changes}
\label{sec:libero_results}

World models used for action selection must accurately predict
the consequences of alternative actions; we evaluate this
in LIBERO~\cite{liu2023libero} by executing two action sequences
from the same initial state and comparing predicted and
reference differences in motion and appearance.
We evaluate 48-step predictions (2.40\,s) by chaining four
12-step segments, conditioning each subsequent segment on
the final predicted frame of the preceding segment within
the same branch.

\input{sec/table_libero_action_response}

ActTrail more accurately reproduces the differences in
appearance and gripper motion between the two reference videos,
as reflected by lower RGB and 2D gripper response errors
(\cref{tab:libero_action_response}).
The reductions in gripper ADE, FDE, and LPIPS also indicate
that the motion and appearance within each predicted branch
more closely match its corresponding reference video.

\subsection{Real-robot video prediction}
\label{sec:realrobot_results}

\input{sec/table_realrobot_results}

We compare DreamDojo and ActTrail on 52 recorded trials
to assess motion fidelity and task-outcome judgments
over 36-step predictions.

ActTrail achieves lower mean frame-to-frame displacement,
endpoint, and dynamic-region RGB $L_1$ errors across all
three tasks (\cref{tab:realrobot_results}), indicating closer
agreement with recorded motion and appearance.
Human evaluation also shows higher agreement with recorded
task outcomes for ActTrail on tube grasping and pipette-tip
attachment, while the two methods perform equally on
tube placement.

\begin{figure}[t]
    \centering
    \includegraphics[width=\columnwidth]
        {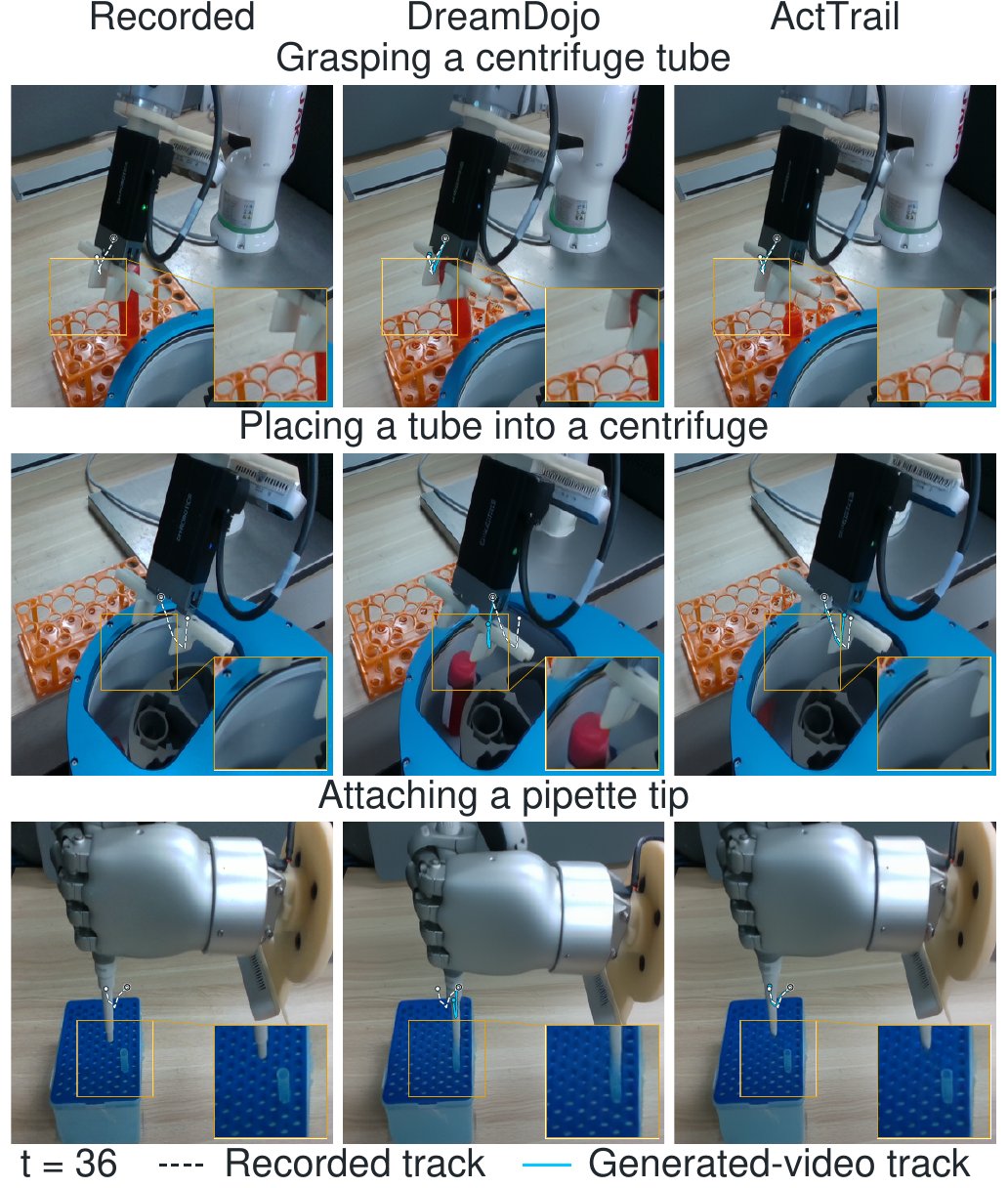}
    \caption{\textbf{Real-robot video prediction at step 36.}
    Rows show tube grasping, tube placement, and pipette-tip
    attachment.
    Columns compare recorded videos, DreamDojo, and ActTrail.
    Magnified views highlight local details in the interaction
    regions.
    Dashed lines indicate recorded tracks, and cyan lines
    indicate tracks in generated videos.}
    \label{fig:realrobot_qualitative}
\end{figure}

In \cref{fig:realrobot_qualitative}, ActTrail reproduces the recorded task outcomes for tube grasping and pipette-tip attachment.
In the tube-placement example, DreamDojo generates an object
at a location inconsistent with the recording, illustrating
that agreement in task outcome does not ensure accurate
local spatial relationships.

\subsection{Ablation studies}
\label{sec:ablation_studies}

\input{sec/table_stage1_ablation}

\noindent\textbf{Trajectory prediction in Stage~1.}
We evaluate Stage~1 on 64 BridgeData V2 development clips
after 30,000 training updates.
Under the same training data and joint objective, the DINOv2
configuration achieves lower cumulative and frame-to-frame
displacement errors in moving regions than the VAE configuration
(\cref{tab:stage1_ablation}), motivating its use for
trajectory prediction.

A second comparison fixes the DINOv2 architecture, initialization,
and training clips.
The weighted joint trajectory and optical-flow objective
improves frame-to-frame prediction over the unit-weight
trajectory objective.

\noindent\textbf{Trajectory fusion in Stage~2.}
To assess how predicted motion guides video generation,
we compare B-Mod, Feature Warping, and ActTrail using the
same frozen Stage~1 predictor and the same number of video
training updates.
The three designs incorporate identical predicted trajectories
through coordinate-based modulation, feature warping, and
path-based feature interaction with spatial write-back,
respectively (\cref{fig:stage2_fusion}).
On the BridgeData V2 test set, ActTrail achieves the lowest
frame-to-frame displacement, endpoint, and dynamic-region
RGB $L_1$ errors (\cref{tab:stage2_fusion}), supporting the
use of predicted paths to organize feature interaction
and spatial write-back.

\input{sec/table_stage2_fusion}

\noindent\textbf{Feature processing within temporal slots.}
Each future latent slot represents four action steps.
We examine whether phase-specific affine parameters and
learned aggregation weights improve motion prediction.
PShared retains all 13 trajectory nodes and spatial write-back
at all 12 future steps, but shares the affine parameters across
the four within-slot phases and uses fixed aggregation weights
of $1/4$.
Full uses phase-specific affine parameters and learned
aggregation weights, both shared across future slots,
and achieves lower mean frame-to-frame displacement and
endpoint errors (\cref{tab:stage2_components}).

\noindent\textbf{Two-stage training.}
We compare our two-stage strategy with Joint.
Joint omits separate Stage~1 pretraining and updates the
trajectory predictor with both motion and video-generation
gradients while jointly optimizing the video model.
It retains the same pretrained DreamDojo backbone and
frozen DINOv2 and VAE as Full.
With the same number of video model updates, staged training
achieves lower motion and dynamic-region reconstruction
errors (\cref{tab:stage2_components}).

\input{sec/table_stage2_components}

%% file: sec/table1_bridge.tex

\begin{table*}[t]
  \centering
  \caption{\textbf{12-step video prediction on BridgeData V2
  and LIBERO.}
  Motion metrics use points jointly valid in the reference
  and all predictions at the required time steps.
  Bold indicates the best mean within each dataset.}
  \label{tab:bridge_main}

  \small
  \setlength{\tabcolsep}{3pt}
  \renewcommand{\arraystretch}{1.0}

  \begin{tabular*}{\textwidth}
    {@{\extracolsep{\fill}}ll*{8}{c}@{}}
    \toprule

    \multirow{2}{*}{Dataset}
    & \multirow{2}{*}{Method}
    & \multicolumn{3}{c}{Motion error $\downarrow$}
    & \multicolumn{2}{c}{Regional error $\downarrow$}
    & \multicolumn{3}{c}{Full-frame quality} \\
    \cmidrule(lr){3-5}
    \cmidrule(lr){6-7}
    \cmidrule(l){8-10}

    &
    & Step
    & Mean cum.
    & End
    & Dyn.~$L_1$
    & Boundary~$L_1$
    & PSNR $\uparrow$
    & SSIM $\uparrow$
    & LPIPS $\downarrow$ \\
    \midrule

    \multirow{5}{*}{\shortstack[l]{BridgeData\\V2}}
    & DreamDojo
    & 12.43 & 30.70 & 55.93
    & 39.43 & 43.55
    & 20.45 & 0.796 & 0.138 \\

    & IRASim
    & 10.80 & 25.84 & 40.77
    & 33.34 & 39.46
    & 19.89 & 0.709 & 0.298 \\

    & Vid2World
    & \textbf{8.13} & 22.22 & 39.64
    & 33.52 & 39.26
    & 20.71 & 0.742 & 0.174 \\

    & Ctrl-World
    & 12.44 & 31.54 & 59.16
    & 39.84 & 44.17
    & 19.63 & 0.759 & 0.176 \\

    & ActTrail
    & 10.01 & \textbf{17.67} & \textbf{27.11}
    & \textbf{29.90} & \textbf{38.71}
    & \textbf{21.39} & \textbf{0.804} & \textbf{0.124} \\

    \addlinespace[1pt]
    & \textit{Valid clips}
    & 120 & 120 & 75
    & 128 & 128
    & 128 & 128 & 128 \\

    \midrule

    \multirow{5}{*}{LIBERO}
    & DreamDojo
    & 0.3046 & 1.2544 & \textbf{1.9741}
    & 14.01 & 19.81
    & 28.83 & 0.9323 & 0.02444 \\

    & IRASim
    & 0.4123 & 1.8296 & 3.1528
    & 17.32 & 24.10
    & 26.50 & 0.8758 & 0.05561 \\

    & Vid2World
    & 0.5140 & 2.5779 & 5.0747
    & 19.21 & 25.83
    & 26.97 & 0.8947 & 0.06071 \\

    & Ctrl-World
    & 0.4929 & 2.5612 & 4.3761
    & 21.57 & 29.01
    & 25.60 & 0.8776 & 0.06053 \\

    & ActTrail
    & \textbf{0.2988} & \textbf{1.2449} & 2.0269
    & \textbf{13.99} & \textbf{19.71}
    & \textbf{28.97} & \textbf{0.9338} & \textbf{0.02383} \\

    \addlinespace[1pt]
    & \textit{Valid clips}
    & 791 & 791 & 775
    & 758 & 778
    & 800 & 800 & 800 \\

    \bottomrule
  \end{tabular*}
\end{table*}

%% file: sec/table_libero_action_response.tex

\begin{table*}[t]
  \centering
  \caption{\textbf{48-step paired-action video prediction on LIBERO.}
  Evaluation uses frames $8,12,\ldots,48$.
  RGB response uses the 0--255 scale;
  gripper errors are measured in image pixels.
  Bold indicates the best mean.}
  \label{tab:libero_action_response}

  \small
  \setlength{\tabcolsep}{3pt}

  \begin{tabular*}{\textwidth}
    {@{\extracolsep{\fill}}l*{6}{c}@{}}
    \toprule
    Method
    & Acc@2 (\%) $\uparrow$
    & RGB response $\downarrow$
    & Gripper response $\downarrow$
    & Gripper ADE $\downarrow$
    & Gripper FDE $\downarrow$
    & LPIPS-Alex $\downarrow$ \\
    \midrule

    DreamDojo
    & 93.125
    & 9.230
    & 6.422
    & 5.621
    & 6.771
    & 0.08327 \\

    IRASim
    & 93.125
    & 10.402
    & 6.777
    & 6.000
    & 7.117
    & 0.13485 \\

    Vid2World
    & \textbf{94.375}
    & 9.656
    & 8.611
    & 5.941
    & 7.197
    & 0.12052 \\

    Ctrl-World
    & 91.250
    & 10.727
    & 8.226
    & 6.377
    & 8.349
    & 0.12579 \\

    ActTrail
    & 93.125
    & \textbf{8.888}
    & \textbf{6.262}
    & \textbf{4.944}
    & \textbf{5.893}
    & \textbf{0.08065} \\

    \midrule
    \textit{Valid initial states}
    & 70 & 70 & 64 & 64 & 70 & 70 \\
    \bottomrule
  \end{tabular*}
\end{table*}

%% file: sec/table_realrobot_results.tex

\begin{table}[t]
    \centering
    \caption{\textbf{36-step real-robot video prediction across three tasks.}
    Bold indicates the best result within each task.}
    \label{tab:realrobot_results}

    \small
    \setlength{\tabcolsep}{3pt}

    \begin{tabular*}{\columnwidth}
        {@{\extracolsep{\fill}}lrrrr@{}}
        \toprule
        Method
        & Step $\downarrow$
        & End $\downarrow$
        & Dyn.~$L_1$ $\downarrow$
        & Acc. (\%) $\uparrow$ \\
        \midrule

        \multicolumn{5}{@{}l}
        {\textit{Grasping a centrifuge tube (20 trials)}} \\

        DreamDojo
        & 0.451
        & 3.242
        & 13.16
        & 85.0 \\

        ActTrail
        & \textbf{0.362}
        & \textbf{1.596}
        & \textbf{9.20}
        & \textbf{95.0} \\

        \midrule
        \multicolumn{5}{@{}l}
        {\textit{Placing a tube into a centrifuge (16 trials)}} \\

        DreamDojo
        & 0.620
        & 7.747
        & 15.29
        & \textbf{100.0} \\

        ActTrail
        & \textbf{0.450}
        & \textbf{2.345}
        & \textbf{8.08}
        & \textbf{100.0} \\

        \midrule
        \multicolumn{5}{@{}l}
        {\textit{Attaching a pipette tip (16 trials)}} \\

        DreamDojo
        & 1.601
        & 20.573
        & 24.25
        & 62.5 \\

        ActTrail
        & \textbf{0.900}
        & \textbf{3.165}
        & \textbf{8.15}
        & \textbf{87.5} \\

        \bottomrule
    \end{tabular*}
\end{table}

%% file: sec/table_stage1_ablation.tex

\begin{table}[t]
    \centering
    \caption{\textbf{Stage~1 trajectory prediction on BridgeData V2.}
    Mean errors over 12 future steps in moving regions,
    measured on the $60\times80$ trajectory grid.}
    \label{tab:stage1_ablation}

    \small
    \setlength{\tabcolsep}{2pt}

    \begin{tabular*}{\columnwidth}
        {@{\extracolsep{\fill}}llrr@{}}
        \toprule
        Features & Objective
        & Mean cum. $\downarrow$
        & Step $\downarrow$ \\
        & & (grid px) & (grid px/interval) \\
        \midrule

        VAE
        & $L_{\mathrm{flow}} + 0.1 L_P$
        & 3.807
        & 1.491 \\

        DINOv2
        & $L_P$
        & 3.783
        & 1.516 \\

        DINOv2
        & $L_{\mathrm{flow}} + 0.1 L_P$
        & \textbf{3.683}
        & \textbf{1.426} \\

        \bottomrule
    \end{tabular*}
\end{table}

%% file: sec/table_stage2_fusion.tex

\begin{table}[t]
    \centering
    \caption{\textbf{Stage~2 trajectory fusion on BridgeData V2.}
    Displacement errors use common valid tracks.}
    \label{tab:stage2_fusion}

    \small
    \setlength{\tabcolsep}{3pt}

    \begin{tabular*}{\columnwidth}
        {@{\extracolsep{\fill}}lccc@{}}
        \toprule
        Fusion
        & Step $\downarrow$
        & End $\downarrow$
        & Dyn.~$L_1$ $\downarrow$ \\
        \midrule

        B-Mod
        & 12.42
        & 42.69
        & 31.26 \\

        Feature Warping
        & 10.93
        & 36.25
        & 30.69 \\

        ActTrail
        & \textbf{10.87}
        & \textbf{35.16}
        & \textbf{29.90} \\

        \bottomrule
    \end{tabular*}
\end{table}

%% file: sec/table_stage2_components.tex

\begin{table}[t]
    \centering
    \caption{\textbf{Stage~2 ablations on BridgeData V2.}
    Each comparison uses its own common valid tracks.}
    \label{tab:stage2_components}

    \small
    \setlength{\tabcolsep}{3pt}

    \begin{tabular*}{\columnwidth}
        {@{\extracolsep{\fill}}lrrrr@{}}
        \toprule
        & \multicolumn{2}{c}{Phase sharing}
        & \multicolumn{2}{c}{Joint training} \\
        \cmidrule(lr){2-3}
        \cmidrule(lr){4-5}
        Metric & Full & PShared & Full & Joint \\
        \midrule

        Step $\downarrow$
        & \textbf{10.84}
        & 11.07
        & \textbf{10.74}
        & 13.14 \\

        End $\downarrow$
        & \textbf{37.05}
        & 38.92
        & \textbf{36.62}
        & 52.29 \\

        Dyn.~$L_1$ $\downarrow$
        & 29.90
        & \textbf{29.62}
        & \textbf{29.90}
        & 35.82 \\

        \bottomrule
    \end{tabular*}
\end{table}

%% file: sec/5_conclusion.tex
\section{Conclusion}
\label{sec:conclusion}

We present ActTrail, which connects robot actions to future
video generation through predicted dense 2D point trajectories.
Point-track and adjacent-frame optical-flow supervision guide
the prediction of cumulative displacement and stepwise motion,
while a trajectory fusion module preserves action-step positions
and cross-step correspondences in the temporally compressed
video model.
Experiments on BridgeData V2, LIBERO, and real robots
demonstrate improvements in motion prediction accuracy
and visual quality in dynamic regions.
Paired-action evaluation further shows that ActTrail more
accurately predicts differences in appearance and gripper
motion between alternative action sequences from the same
initial state.
These results support predicted point trajectories as an
effective intermediate motion representation for more
accurately depicting scene changes induced by robot actions.

%% file: sec/6_llm_usage.tex
\section*{LLM Usage Statement}

During the research and preparation of this paper, we used
ChatGPT (GPT-series models) and Codex, both provided by OpenAI.
These tools were used to polish the language and improve the
clarity of text written by the authors; assist with organizing
related literature and drafting portions of the Related Work
section; generate and modify portions of the implementation
code; and suggest revisions to the authors' experimental plans,
analyses of results, figure presentation, and scientific
arguments.
These tools were not used for synthetic data generation,
dataset construction, or data annotation.

The authors are responsible for reviewing, verifying, and
revising all tool outputs adopted in this work.
They take full responsibility for the accuracy and completeness
of the final submission, including its references, code,
experimental results, and scientific conclusions.